\documentclass[runningheads]{llncs}
\usepackage{graphicx}
\usepackage{color}
\usepackage{times}
\usepackage{epsfig}
\graphicspath{{./imgs/}}
\usepackage{verbatim}
\usepackage{booktabs}
\usepackage[hidelinks=true]{hyperref}
\hypersetup{
	colorlinks   = true, %Colours links instead of ugly boxes
	urlcolor     = green, %Colour for external hyperlinks
	linkcolor    = red, %Colour of internal links
	citecolor   = green %Colour of citations
}

\begin{document}
%===========================================================

\title{Learning to Clean: A GAN Perspective} % Replace your paper's title here
\titlerunning{Document Cleaning Suite} % Replace an abstracted version of your paper's title here

%===========================================================

\author{Monika Sharma \and
Abhishek Verma \and
Lovekesh Vig}
%
%Please include author names in full in the paper, 
%If any authors have names that can be parsed into FirstName LastName in multiple ways, please include the correct parsing, in a comment to the volume editors:
%\index{Lastnames, Firstnames}

\authorrunning{M. Sharma et al.} % A shorter version of authors' name

%===========================================================

\institute{TCS Research, New Delhi, India\\
\email{\{monika.sharma1 $\mid$ verma.abhishek7 $\mid$ lovekesh.vig \}@tcs.com}\\
}

\maketitle

%===========================================================
\begin{abstract}
In the big data era, the impetus to digitize the vast reservoirs of data trapped in unstructured scanned documents such as invoices, bank documents, courier receipts and contracts has gained fresh momentum. The scanning process often results in the introduction of artifacts such as salt-and-pepper / background noise, blur due to camera motion or shake, watermarkings, coffee stains, wrinkles, or faded text. These artifacts pose many readability challenges to current text recognition algorithms and significantly degrade their performance. Existing learning based denoising techniques require a dataset comprising of noisy documents paired with cleaned versions of the same document. In such scenarios, a model can be trained to generate clean documents from noisy versions. However, very often in the real world such a paired dataset is not available, and all we have for training our denoising model are unpaired sets of noisy and clean images. This paper explores the use of Generative Adversarial Networks (GAN) to generate denoised versions of the noisy documents. In particular, where paired information is available, we formulate the problem as an image-to-image translation task i.e, translating a document from noisy domain ( i.e., background noise, blurred, faded, watermarked ) to a target clean document using Generative Adversarial Networks (GAN). However, in the absence of paired images for training, we employed CycleGAN which is known to learn a mapping between the distributions of the noisy images to the denoised images using unpaired data to achieve image-to-image translation for cleaning the noisy documents. We compare the performance of CycleGAN for document cleaning tasks using unpaired images with a Conditional GAN trained on paired data from the same dataset. Experiments were performed on a public document dataset on which different types of noise were artificially induced, results demonstrate that CycleGAN learns a more robust mapping from the space of noisy to clean documents. 

\end{abstract}
%===========================================================

\section{Introduction}
\label{sec:intro}
The advent of industry 4.0 calls for the digitization of every aspect of industry, which includes automation of business processes, business analytics and phasing out of manually driven processes. While business processes have evolved to store large volumes of scanned digital copies of paper documents, however for many such documents the information stored needs to be extracted via text recognition techniques. While capturing these images via camera or scanner, artifacts tend to creep into the images such as background noise, blurred and faded text. In some scenarios, companies insert a watermark in the documents which poses readability issues after scanning. Text recognition engines often suffer due to the low quality of scanned documents and are not able to read the documents properly and hence, fail to correctly digitize the information present in the documents. In this paper, we attempt to perform denoising of the documents before the document is being sent to text recognition network for reading and propose a document cleaning suite based on generative adversarial training. This suite is trained for background noise removal, deblurring, watermark removal and defading and learns a mapping from the distribution of noisy documents to the distribution of clean documents.

Background noise removal is the process of removing the background noise, such as uneven contrast, see through effects, interfering strokes, and background spots on the documents. The background noise presents a problem to the performance of OCR as it is difficult to differentiate the text and background~\cite{denoise_1},~\cite{denoise_2},~\cite{denoise_3},~\cite{denoise_4}. De-blurring is the process of removal of blur from an image. Blur is defined as distortion in the image due to various factors such as shaking of camera, improper focus of camera etc. which decreases the readability of the text in the document image and hence, deteriorates the performance of OCR. Recent works for deblurring have focused on estimating blur kernels using techniques such as GAN~\cite{gan}, CNN~\cite{cnn_deblur}, dictionary-based prior~\cite{deblur_dictionary}, sparsity-inducing prior~\cite{sparsity_deblur} and hybrid non-convex regularizer~\cite{deblur_hybrid}. Watermark removal aims at removing the watermark from an image while preserving the text in the image. Watermarks are low-intensity images printed on photographs and books in order to prevent copying of the material. But this watermark post scanning creates hinderance in reading the text of interest from documents. Inpainting~\cite{watermark_inpainting1}~\cite{watermark_inpainting2} techniques are used in the literature to recover the original image after detecting watermarks statistically. Defading is the process of recovering text that has lightened / faded over time, which usually happens in old books and documents. This is also detrimental to the OCR performance. To remove all these artifacts that degrade the quality of documents and create hindrance in readability, we formulate the document cleaning process as an image-to-image translation task at which Generative Adversarial Networks (GANs)~\cite{gan} are known to give excellent performance. 

However, with the limited availability of paired data i.e., noisy and corresponding cleaned documents, we proposed to train CycleGAN~\cite{cyclegan} for unpaired datasets of noisy documents. We train CycleGAN for denoising / background noise removal, deblurring, watermark removal and defading tasks. CycleGAN eliminates the need for one-to-one mapping between images of source and target domains by a two-step transformation of source image i.e., first source image is mapped to an image in target domain and then back to source again. We evaluate the performance of our document cleaning suite on synthetic and publicly available datasets and compare them against state-of-the-art methods. We use Kaggle's document dataset for denoising / background noise removal~\cite{kaggle_dataset}, the BMVC document deblurring dataset~\cite{bmvc_dataset} which are publicly available online. There does not exist any document dataset for watermark removal and defading online. Therefore, we have synthetically generated document datasets for watermark removal and defading tasks, and have also made these public for the benefit of research community. Overall, our contributions in this paper are as follows :
\begin{itemize}
\item We proposed a Document Cleaning Suite which is capable of cleaning documents via denoising / background noise removal, deblurring, watermark removal and defading for improving readability.
\item We proposed the application of CycleGAN~\cite{cyclegan} for translating a document from a noisy document distribution (e.g. with background noise, blurred, watermarked and faded) to a clean document distribution in the situations where there is shortage of paired dataset.
\item We synthetically created a document dataset for watermark removal and defading by inserting logos as watermarks and applying fading techniques on Google News dataset~\cite{google_dataset} of documents, respectively. %We also make these datasets~\cite{} public for evaluation by third party.
\item We evaluate CycleGAN for background noise removal, deblurring, watermark removal and defading on publicly available kaggle document dataset~\cite{kaggle_dataset}, BMVC deblurring document dataset~\cite{bmvc_dataset} and synthetically created watermarked and defading document datasets, respectively.
\end{itemize}

The remaining parts of the paper are organized as follows. Section~\ref{sec:related} reviews the related work. Section~\ref{sec:cycle-gan} introduces CycleGAN and explains its architecture. Section~\ref{sec:results-discussions} provides details of datasets, training, evaluation metric used and also discusses experimental results and comparisons to evaluate the effectiveness and superiority of CycleGAN for cleaning the noisy documents. Section~\ref{sec:conclusion} concludes the paper.

\section{Related Work}
\label{sec:related}
	
Generative adversarial Network (GAN)~\cite{gan} is the idea that has taken deep learning by storm. It employs adversarial training which essentially means pitting two neural networks against each other. One is a generator while the other is a discriminator, where the former aims at producing data that are indistinguishable from real data while the latter tries to distinguish between real and fake data. The process eventually yields a generator with the ability to do a plethora of tasks efficiently such as image-to-image generation. Other notable applications where GANs have established their supermacy are representation learning, image editing, art generation, music generation etc.~\cite{infogan}~\cite{cm_gan}~\cite{pm_gan}~\cite{marta_gan}~\cite{dc_gan}.

Image-to-image translation is the task of mapping images in source domain to images in target domain such as converting sketches into photographs, grayscale images to color images etc. The aim is to generate the target distribution given the source distribution. Prior work in the field of GANs such as Conditional GAN~\cite{conditional_gan} forces the image produced by generator to be conditioned on the output which allows for optimal translations. However, earlier GANs require one-to-one mapping of images between source and target domain i.e., a paired dataset. In case of documents, it is not possible to always have cleaned documents corresponding to each noisy document. This persuaded us to explore unpaired image-to-image translation methods, e.g. Dual-GAN~\cite{dual_gan} which uses dual learning and CycleGAN~\cite{cyclegan} which makes use of cyclic-consistency loss to achieve unpaired image-to-image translation.

In this paper, we propose to apply CycleGAN for document cleaning task. It has two pairs of generators and discriminators. One pair focuses on converting source domain to target domain while the other pair focuses on converting target domain to source domain. This bi-directional conversion process allows for a cyclic consistency loss for CycleGAN which ensures the effective conversion of an image from source to target and then back to source again. The transitivity property of cyclic-consistency loss allows CycleGAN to perform well on unpaired image-to-image translation.

Existing methods for removing background noise from document images consist of binarization and thresholding techniques, fuzzy logic, histogram, morphology and genetic algorithm based methods~\cite{denoise_1}~\cite{denoise_2}. An automatic method for color noise estimation from a single image using Noise Level Function (NLF) and a Gaussian Conditional Random Field (GCRF) based removal technique was proposed in ~\cite{denoise_3} for producing a clean image from noisy input. Sobia et al.~\cite{denoise_4} employs a technique for removing background and punch-hole noise from handwritten Urdu text. We observed that deep learning has not been applied in literature for removing noise from document images.
	
There exists quite a lot of work on deblurring of images. For example, DeblurGAN~\cite{deblur_gan} uses conditional GANs to deblur images, ~\cite{multicnn_deblur} uses a multi-scale CNN to create an end-to-end system for deblurring. Ljubenovic et al. proposed class-adapted dictionary-based prior for the image~\cite{classadapted_deblur}. There also exists method of sparsity-inducing prior on the blurring filter, which allows for deblurring images containing different classes of images such as faces, text etc.~\cite{sparsity_deblur} when they co-occur in a document. A non-convex regularization method was developed by Yao et al.~\cite{deblur_hybrid} which leveraged the non-convex sparsity constraints on image gradients and blur kernels for improving the kernel estimation accuracy. ~\cite{cnn_deblur} uses a CNN to classify the image into one of the degradative sub-spaces and the corresponding blur kernel is then used for deblurring.

Very few attempts have been made in past for removing watermarks from images. Authors in ~\cite{watermark_inpainting1} proposed to use image inpainting to recover the original image. However, the method developed by Xu et al.~\cite{watermark_inpainting2} detects the watermark using statistical methods and subsequently, removes it using image inpainting. To the best of our knowledge, we did not find any work on defading of images.

	%===========================================================
\begin{figure}[h]
\begin{center}
   \includegraphics[width=0.9\linewidth]{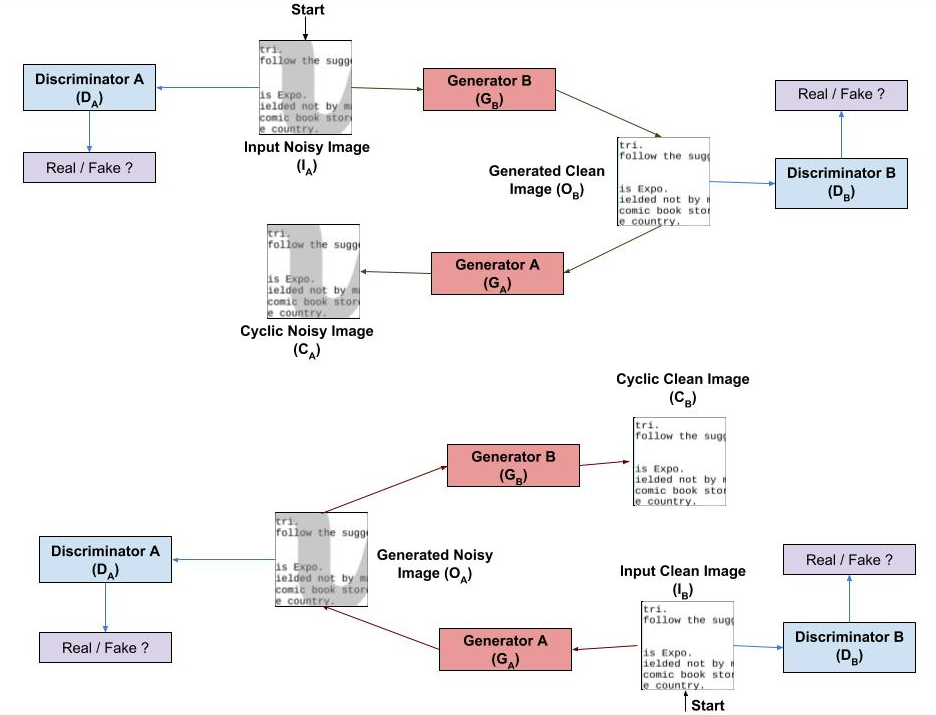}
\end{center}
   \caption{Overview of CycleGAN - It consists of two generators, $G_A$ and $G_B$ which map noisy images to clean images and clean to noisy images, respectively using cycle-consistency loss~\cite{cyclegan}. It also contains two discriminators $D_A$ and $D_B$ which acts as adversary and rejects images generated by generators.}
\label{fig:cyclegan-overview}
%\label{fig:onecol}
\end{figure}

\section{CycleGAN}
\label{sec:cycle-gan}
%\\\\\\\\\\\\\\\\\\\\\\\\\\\\\\\\\\\\\\\\\\\\\\\\\\\\
%Our goal is obtain clean images in case when the images are blurred, watermarked, faded and have background in an unpaired setting. For this, we use Cycle GAN. The aim is take both sets of images and convert one into another. To achieve this, two pairs of generator-discriminator are used, one that handles dirty to clean transformation and the other handles clean to dirty formulation. But, this can lead to a problem that input when transformed from one distribution to another can map to many images. In order to alleviate this, we make sure that transformation of dirty to clean leads back to the original dirty image and vice-versa also remains true. This is the aim of cyclic consistency loss.

CycleGAN~\cite{cyclegan} has shown its worth in scenarios where there is paucity of paired dataset, i.e., image in source domain and corresponding image in target domain. This property of CycleGAN, of working without the need of one-to-one mapping between input domain and target domain and still being able to learn such image-to-image translations, persuades us to use them for document cleaning suite where there is always limited availability of clean documents corresponding to noisy documents. To circumvent the issue of learning meaningful transformations in case of unpaired dataset, CycleGAN uses cycle-consistency loss which says that if an image is transformed from source distribution to target distribution and back again to source distribution, then we should get samples from source distribution. This loss is incorporated in CycleGAN by using two generators and two discriminators, as shown in Figure~\ref{fig:cyclegan-overview}. The first generator $G_{B}$ maps the image from noisy domain A ($I_{A}$) to an output image in target clean domain B ($O_{B}$). To make sure that there exists a meaningful relation between $I_{A}$ and $O_{B}$, they must learn some features which can be used to map back $O_{B}$ to original noisy input domain. This reverse transformation is carried out by second generator $G_{A}$ which takes as input $O_{B}$ and converts it back into an image $C_{A}$ in noisy domain. Similar process of transformation is carried out for converting images in clean domain B to noise domain A as well. It is evident in the Figure~\ref{fig:cyclegan-overview} that each discriminator takes two inputs - original image in source domain and generated image via a generator. The task of the discriminator is to distinguish between them so that discriminator is able to defeat generator by rejecting images generated by it. While competing against discriminator so that it stops rejecting its images, the generator learns to produce images very close to the original input images.

	%==========================================================
We use the same network of CycleGAN as proposed in ~\cite{cyclegan}. The generator network consists of two convolutional layers of stride 2, several residual blocks, two layers of transposed convolutions with stride 1. The discriminator network uses $70 \times 70$ PatchGANs~\cite{patchgan} to classify the $70 \times 70$ overlapping patches of images as real or fake.

	%------------------------------------------------------------------------- 
\section{Experimental Results and discussion}
\label{sec:results-discussions}
This section is divided into the following subsections: Section~\ref{subsec:dataset} provides details of the datasets used for the document cleaning suite. In Section~\ref{subsec:training-details}, we elaborate on the training details utilized to perform our experiments. Next, we give the performance evaluation metric in Section~\ref{subsec:eval-metric}. Subsequently, Section~\ref{subsec:results} discusses the results obtained from the experiments we conducted and provides comparison with the baseline model i.e., Conditional GAN~\cite{conditional_gan}.

\begin{table}[h]
\caption{Performance comparison of Conditional GAN and CycleGAN based on PSNR}
\centering{
\begin{tabular}{| l | c | c | c | c |}
\hline
& \multicolumn{2}{|c|}{\textbf{PSNR (in dB)}} \\
\hline
\textbf{Task} & \textbf{ConditionalGAN} & \textbf{CycleGAN}\\
\hline
Background removal &  $27.624$  &  $\textbf{31.774}$ \\
\hline
Deblurring &  $19.195$ & $\textbf{30.293}$  \\
\hline
Watermark removal &  $29.736$ & $\textbf{34.404}$  \\
\hline
Defading &  $28.157$ & $\textbf{34.403}$ \\
\hline
\end{tabular}
}\\
\label{tab:results-table}
\end{table}

\begin{figure}[h]
\begin{center}
   \includegraphics[width=0.7\linewidth]{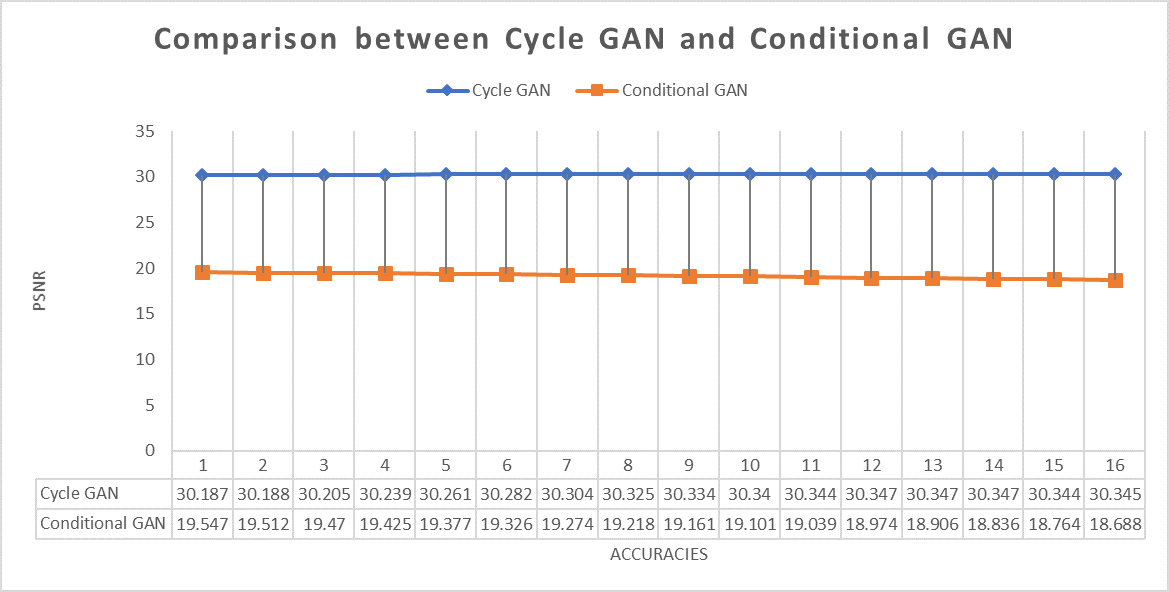}
\end{center}
   \caption{Plot showing comparison between PSNR of images produced by CycleGAN~\cite{cyclegan} and ConditionalGAN~\cite{conditional_gan} on test-set of deblurring document dataset~\cite{bmvc_dataset}. The test-set consists of 16 sets of 100 documents each, where each set is blurred with one of the 16 different blur kernels used for creating the training dataset.}
\label{fig:deblur-psnr}
%\label{fig:onecol}
\end{figure}

\subsection{Dataset Details}
\label{subsec:dataset}
We used 4 separate document datasets, one each for background noise removal, deblurring, watermark removal and defading. Their details are given below :
\begin{itemize}
\item \textbf{Kaggle Document Denoising Dataset} : This document denoising dataset hosted by Kaggle~\cite{kaggle_dataset} consists of noisy documents with noise in various forms such as coffee stains, faded sun spots, dog-eared pages, and lot of wrinkles etc. We use this dataset for training and evaluating CycleGAN for removing background noise from document images. We have used a training set of 144 noisy documents to train CycleGAN and tested the trained network on a test dataset of 72 document images. \\

%CycleGAN was run for 12 epochs on this dataset.\\

\item \textbf{Document Deblurring Dataset} : We used artificial deblurring dataset of documents~\cite{bmvc_dataset} available online for training CycleGANs to deblur the blurred documents. This dataset was created by taking documents from the CiteSeerX repository and were further processed via various geometric transformations and two types of blur i.e., motion blur and de-focus blur, on them to make the noise look more realistic. We have used only a subset of this dataset by random sampling of 2000 documents for training CycleGAN. For evaluation, this deblurring dataset has a test-set which consists of 16 sets of 100 documents, with each set blurred with one of the 16 different blur kernels used for creating the training dataset. \\

%For de-blurring, the dataset used is BMVC dataset. 2000 images each from clean and dirty set were used by random sampling. For test, 200 images were used. Each image is of size 300x300. CycleGAN was run for 60 epochs on this dataset.\\
	
\item \textbf{Watermark Removal Dataset} : As there exists no publicly available dataset for watermarked document images, we generated our own synthetic watermark removal document dataset. To create the dataset, we first obtained text documents from Google News Dataset~\cite{google_dataset} and approx. 21 logos from the Internet for inserting watermarks. Then, we pasted the logos on the documents by making logos transparent with varying values of alpha channel. We used variations in the position of logo, size of logo and transparency factor for creating randomness in the watermarked documents and to make them realistic. The training set of 2000 images and test set of 200 images from this synthetic dataset was used for experimental purposes. \\

%CycleGAN was run for 12 epochs on this dataset.\\
	
\item \textbf{Document Defading Dataset} : Similar to watermark removal dataset, we artificially generated faded documents from Google News Dataset~\cite{google_dataset} by applying various dilation operations on document images. Here again, the train and test set consisted of 2000 and 200 document images, respectively for training and evaluating the performance of CycleGAN for defading purposes.\\
%CycleGAN was run for 8 epochs on this dataset.
\end{itemize}

\subsection{Training Details}
\label{subsec:training-details}
We use the same training procedure as adopted for CycleGan in paper~\cite{cyclegan}. Least-squares loss is used to train the network as this loss is more stable and produces better quality images. We update the discriminators by using a history of generated images rather than the ones produced by latest generator to reduce model oscillations. We use Adam optimizer with learning rate of 0.0002 and momentum of 0.5 for training CycleGAN on noisy images of size $200 \times 200$. The network is trained for 12, 30, 12 and 8 epochs for background noise removal, deblurring, watermark removal and defading, respectively.

For Conditional GAN~\cite{conditional_gan}, we use kernel size of $3\times 3$ with a stride $1$ and zero-padding by $1$ for all convolutional and deconvolutional layers of generator network. In case of discriminator network, the first three convolutional and deconvolutional layers were composed of kernels of size $4\times 4$ with a stride $2$ and zero-padding by $1$. However, the last two layers in discriminator network uses kernel of size $4 \times 4$ with stride of size $1$. The network is trained on input images of size $200 \times 200$  using Adam Optimizer with a learning rate of $2\times 10^{-3}$. We use $6.6\times 10^{-3}$ and $1$ as values of weights for adversarial loss and perceptual loss, respectively. The network is trained for 5 epochs for each of the document cleaning tasks i.e., background noise removal, deblurring, watermark removal and defading.

\subsection{Evaluation Metric}
\label{subsec:eval-metric}
We evaluate the performance of CycleGAN using Peak Signal-to-Noise Ratio (PSNR)\footnote{Peak Signal-to-Noise Ratio: http://www.ni.com/white-paper/13306/en/} as an image quality metric. PSNR is defined as ratio of the maximum possible power of a signal and the power of distorting noise which deteriorates the quality of its representation. PSNR is usually expressed in terms of Mean-squared error (MSE). Given a denoised image (D) of size $m \times n$ and its corresponding noisy image (I) of same size, PSNR is given as follows :
\begin{equation}
PSNR = 20 \times \log10(\frac{Max_{D}}{MSE})
\end{equation}
where $Max_{D}$ represents the maximum pixel intensity value of image D. Higher the PSNR value, better is the image quality.

\begin{figure}[h]
\begin{center}
   \includegraphics[width=0.75\linewidth]{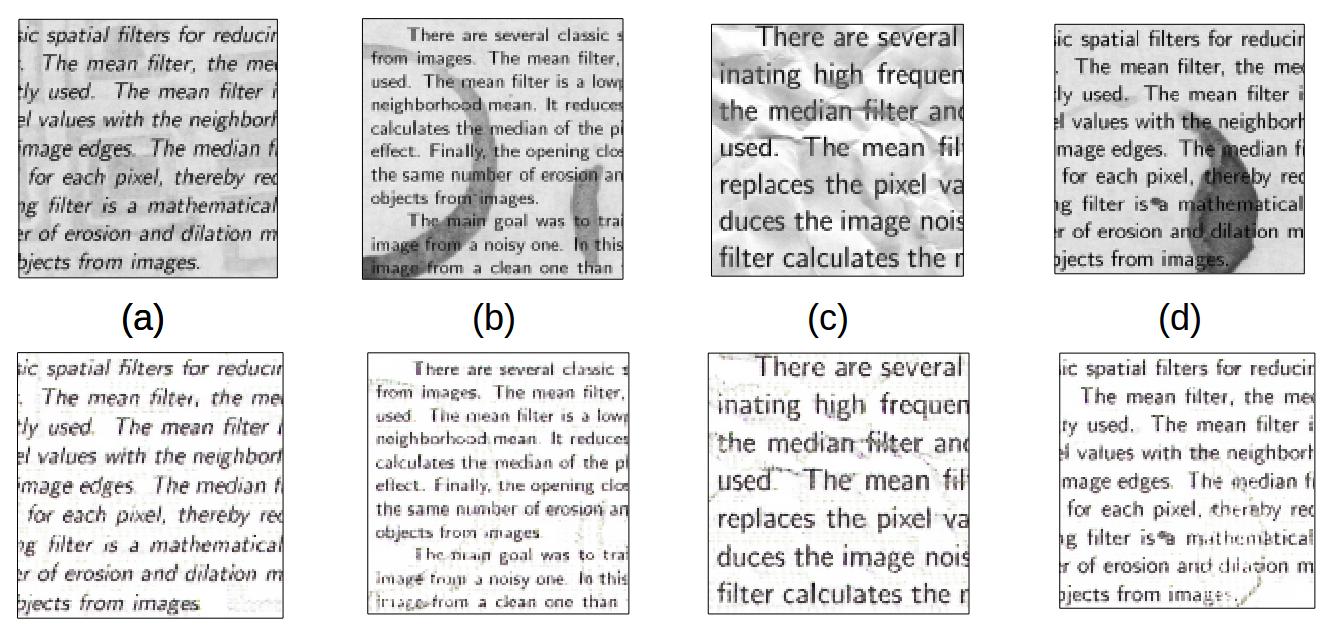}
\end{center}
   \caption{Examples of sample noisy images (upper row) cleaned by CycleGAN and their corresponding cleaned images (bottom row) from Kaggle Document Denoising Dataset~\cite{kaggle_dataset} }
\label{fig:denoising-examples}
%\label{fig:onecol}
\end{figure}

\begin{figure}[h]
\begin{center}
   \includegraphics[width=0.75\linewidth]{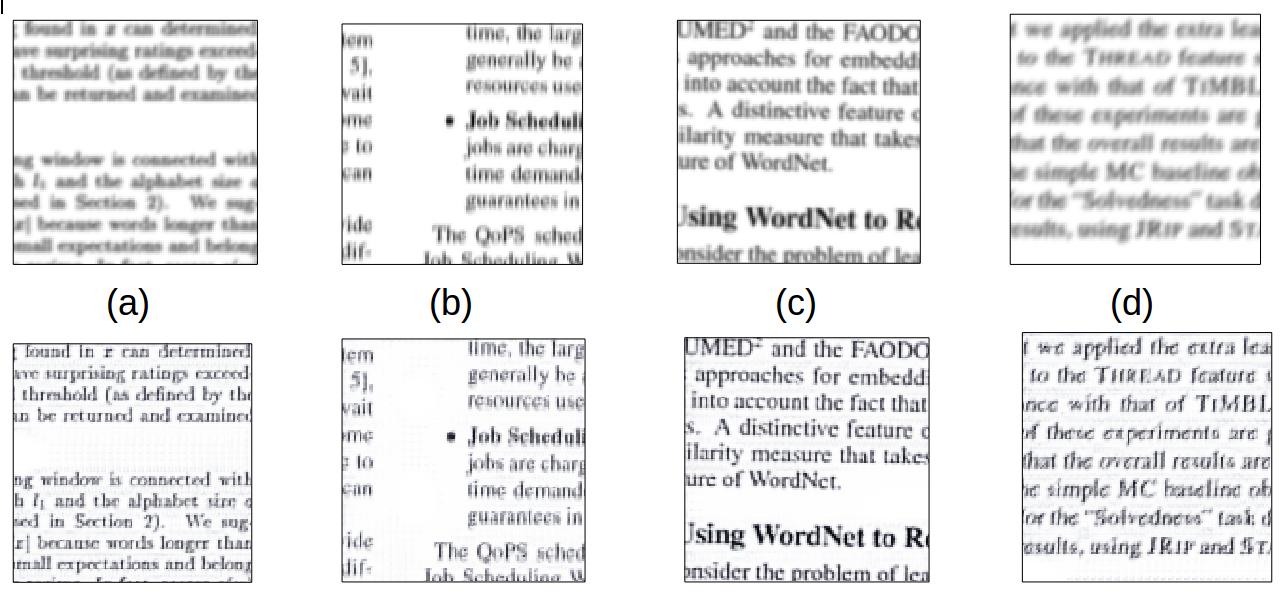}
\end{center}
   \caption{Results of CycleGAN on deblurring document dataset~\cite{bmvc_dataset}. Top row shows the blurred images and bottom row shows their corresponding deblurred images.}
\label{fig:deblur-examples}
%\label{fig:onecol}
\end{figure}

\begin{figure}[h]
\begin{center}
   \includegraphics[width=0.75\linewidth]{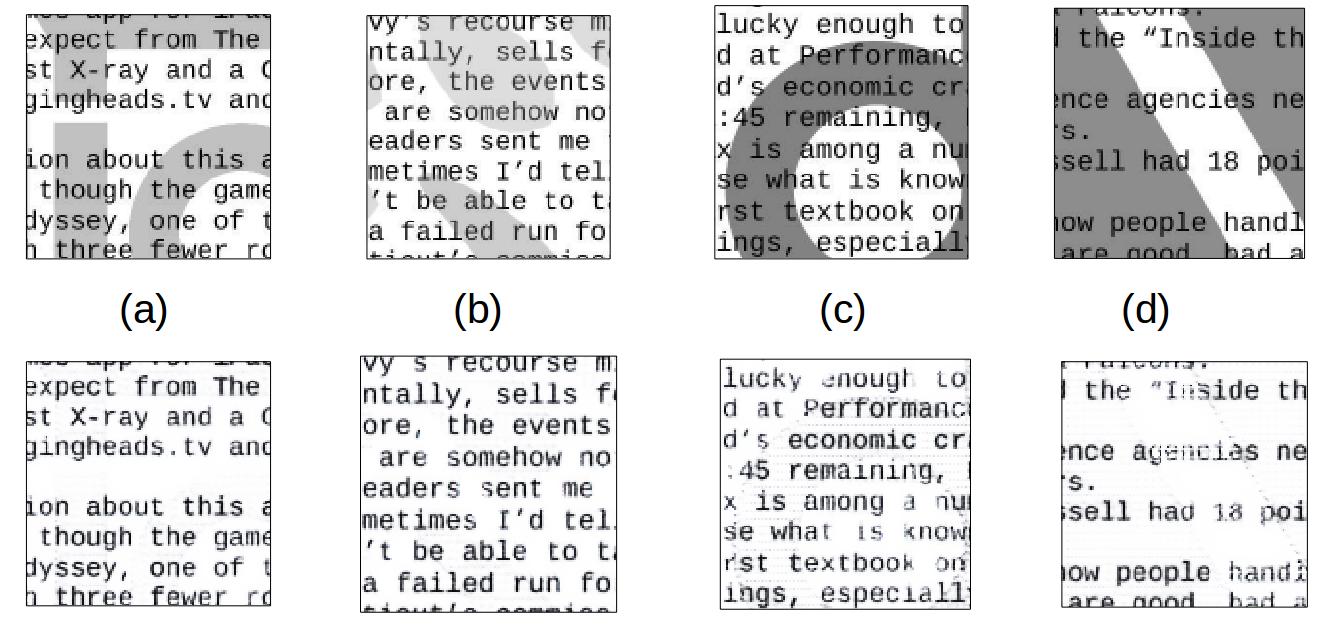}
\end{center}
   \caption{Samples of watermarked images (first row) and their respective cleaned images (second row) produced by CycleGAN.}
\label{fig:watermark-examples}
%\label{fig:onecol}
\end{figure}

\begin{figure}[h]
\begin{center}
   \includegraphics[width=0.75\linewidth]{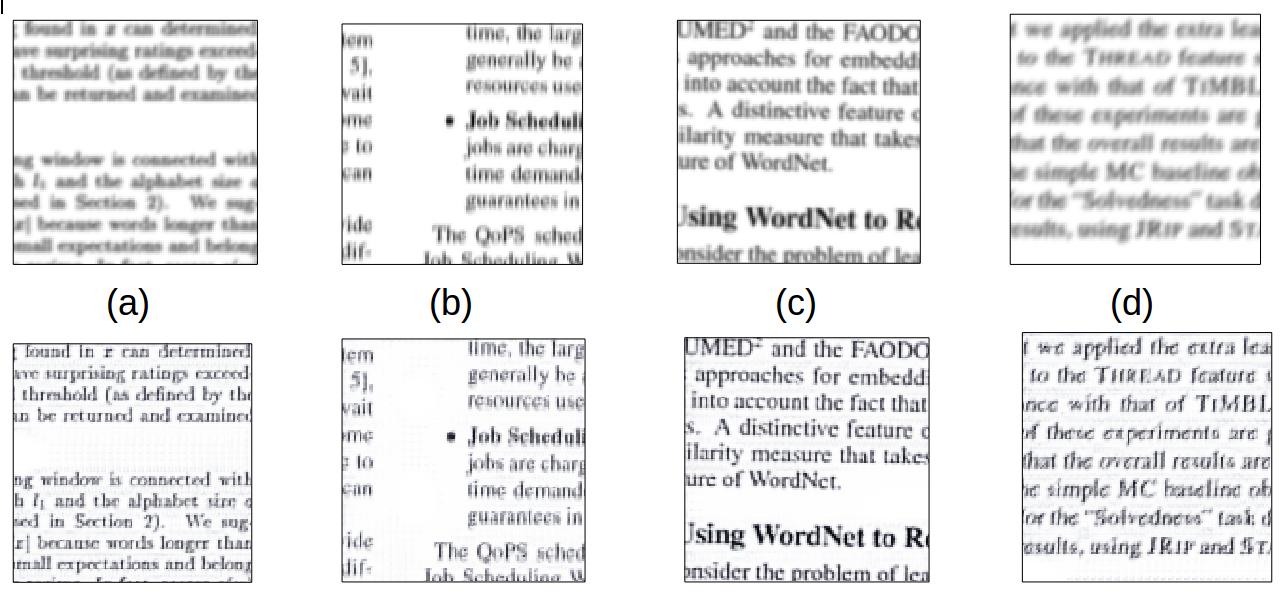}
\end{center}
   \caption{Figure showing example images of faded (top row) and corresponding defaded images (bottom row) with recovered text by CycleGAN.}
\label{fig:defading-examples}
%\label{fig:onecol}
\end{figure}

\subsection{Results}
\label{subsec:results}
Now, we present the results obtained on document datasets using CycleGAN for document cleaning purposes. Table~\ref{tab:results-table} gives the comparison of Conditional GAN and CycleGAN for denoising, deblurring, watermark removal and defading tasks. We observe that CycleGAN beats Conditional GAN on all these document cleaning tasks as shown in Table~\ref{tab:results-table}. Row 1 of Table~\ref{tab:results-table} gives mean PSNR values of images deblurred using Conditional GAN and CycleGAN. CycleGAN obtains higher PSNR value of $31.774$ dB as compared to that of Conditional GAN's PSNR ($27.624$ dB) on Kaggle Document Denoising dataset~\cite{kaggle_dataset}. Similarly, PSNR value of CycleGAN ($19.195$ dB) is better than Conditional GAN for deblurring dataset~\cite{bmvc_dataset}. We have also shown the PSNR comparison for deblurring test-set using a plot, as given in Figure~\ref{fig:deblur-psnr} which shows the superiority of CycleGAN over Conditional GAN. Row 3 and 4 gives the PSNR values for watermark removal and defading task. Here again, CycleGAN gives better image quality.

We also show some sample examples of clean images produced after the application of CycleGAN for all four tasks - background noise removal, deblurring, watermark removal and defading, as given in Figures~\ref{fig:denoising-examples},~\ref{fig:deblur-examples},~\ref{fig:watermark-examples},~\ref{fig:defading-examples}, respectively.

	%------------------------------------------------------------------------ 
\section{Conclusion}
\label{sec:conclusion}
In this paper, we proposed and developed Document Cleaning Suite which is based on the application of CycleGAN and is responsible for performing various document cleaning tasks such as background noise removal, deblurring, watermark removal and defading. Very often it is difficult to obtain clean images corresponding to a noisy image, and simulation of noise for training image-to-image translators does not adequately generalize to the real world. Instead,  we trained a model to learn the mapping from an input distribution to an output distribution of images, while preserving the essence of the image. We used CycleGAN because it has been seen to provide good results for such domain adaptation scenarios where there is limited availability of paired datasets i.e., noisy and correspondig cleaned image. We demonstrated the effectiveness of CycleGAN on publicly available and synthetic document datasets, and the results demonstrate that it can clean up a variety of noise effectively. 
	%===========================================================
\bibliographystyle{splncs04}
\bibliography{accv2018cameraready}
	
	%this would normally be the end of your paper, but you may also have an appendix
	%within the given limit of number of pages
	
\end{document}